\documentclass{article}
\usepackage{kr}

\usepackage{times}
\usepackage{soul}
\usepackage{url}
\usepackage[hidelinks]{hyperref}
\usepackage[utf8]{inputenc}
\usepackage[small]{caption}
\usepackage{graphicx}
\usepackage{amsmath}
\usepackage{amsthm}
\usepackage{booktabs}
\usepackage{algorithm}
\usepackage{algorithmic}
\usepackage{latexsym, amsmath, amssymb, amsfonts, mathrsfs}
\usepackage{xspace}
\usepackage[T1]{fontenc}
\usepackage{float}
\usepackage{color}
\usepackage{url}
\usepackage{todonotes}
\usepackage{picinpar}
\usepackage{multirow}
\usepackage{subcaption}
\usepackage{amsthm}
\usepackage{booktabs} 
\usepackage{diagbox}
\allowdisplaybreaks
\usepackage{pgfplots}

\newcommand{\fml}[1]{\mathcal{#1}}

\newcommand{\beitemize}{\begin{list}{$\bullet$}{\topsep=1.5pt \parsep=0pt \itemsep=1pt \leftmargin=1em }} 
\newcommand{\enitemize}{\end{list}}

\newcommand{\beenumerate}{\hspace{-0.5in} \begin{enumerate}\topsep=1pt \parsep=0pt \itemsep=-3pt} \newcommand{\enenumerate}{\end{enumerate}}

\newcommand{\belist}{\begin{list}{$\bullet$}{\topsep=1.5pt \parsep=0.5pt \itemsep=1pt \leftmargin=2.25em \labelwidth=1.0em \labelsep=0.5em \partopsep=1.5pt}} 
\newcommand{\enlist}{\end{list}}

\newboolean{includeMemo}
\setboolean{includeMemo}{true} 

\newcommand{\memoside}[1]{\ifthenelse{\boolean{includeMemo}}{\todo[caption={},color=green!20!]{{\footnotesize #1}}}}
\newcommand{\memo}[1]{\ifthenelse{\boolean{includeMemo}}{\todo[inline,caption={},color=green!20!]{#1}}}
\newcommand{\memob}[1]{\ifthenelse{\boolean{includeMemo}}{\todo[inline,caption={},color=blue!20!]{#1}}}

\newcommand{\xhdr}[1]{\vspace{5pt}\noindent\textbf{#1 }}
\newcommand{\ignore}[1]{}

\newcommand{\squishlist}{
\begin{list}{{{\small{$\bullet$}}}}
{\setlength{\itemsep}{3pt}      
\setlength{\parsep}{3pt}
\setlength{\topsep}{3pt}       
\setlength{\partopsep}{3pt}
\setlength{\leftmargin}{1em} 
\setlength{\labelwidth}{1em}
\setlength{\labelsep}{0.5em} } }
\newcommand{\squishend}{  \end{list}}

\newcommand{\squishenum}{
\begin{list}{$\bullet$}{ 
    \setlength{\itemsep}{1pt}
    \setlength{\parsep}{0pt}
    \setlength{\topsep}{1.5pt}
    \setlength{\partopsep}{0pt}
    \setlength{\leftmargin}{2em}
    \setlength{\labelwidth}{1.5em}
    \setlength{\labelsep}{0.5em} } }

\title{How Do People Revise Inconsistent Beliefs? \\ Examining Belief Revision in Humans with User Studies}

\author{%
Stylianos Loukas Vasileiou$^1$\and
Antonio Rago$^2$\and
Maria Vanina Martinez$^3$\and
William Yeoh$^1$ \\
\affiliations
$^1$Washington University in St. Louis\\
$^2$Imperial College London\\
$^3$Artificial Intelligence Research Institute (IIIA-CSIC)\\
\emails
\{v.stylianos, wyeoh\}@wustl.edu,
a.rago@imperial.ac.uk,
vmartinez@iiia.csic.es
}

\begin{document}

\maketitle

\begin{abstract}
Understanding how humans revise their beliefs in light of new information is crucial for developing AI systems which can effectively model, and thus align with, human reasoning. While theoretical belief revision frameworks rely on a set of principles that establish how these operations are performed, empirical evidence from cognitive psychology suggests that people may follow different patterns when presented with conflicting information.~In this paper, we present three comprehensive user studies showing that people consistently prefer \emph{explanation-based revisions}, i.e., those which are guided by explanations, that result in changes to their belief systems that are not necessarily captured by classical belief change theory. Our experiments systematically investigate how people revise their beliefs with explanations for inconsistencies, whether they are provided with them or left to formulate them themselves, demonstrating a robust preference for what may seem non-minimal revisions across different types of scenarios. These findings have implications for AI systems designed to model human reasoning or interact with humans, suggesting that such systems should accommodate explanation-based, potentially non-minimal belief revision operators to better align with human cognitive processes.

\end{abstract}

\section{Introduction}

When faced with new information that challenges their existing beliefs, 
people naturally seek to adjust these beliefs to accommodate this information. But how do people actually make these adjustments? Understanding this process is particularly crucial for AI systems that need to collaborate effectively with humans and maintain accurate models of their beliefs and understanding.

Generally, in the field of belief revision (BR)~\cite{alchourron1985logic}, such operations are theoretically defined by a series of postulates or axioms that characterize their behavior. These postulates govern the end result of the operation without making explicit how they are effectively constructed. 
Some of these postulates, such as success and recovery \cite{levi1991fixation,makinson1997force,rott2000two,hansson1999survey,FERME2024109108}, have been challenged in the literature, which prompted the development of alternative postulates and operators.
Another controversial perspective on BR theory is the principle of \textit{minimalism} (or information economy).
While different operators provide alternative perspectives on what minimalism means~\cite{HERZIG1999107}, at its core assumes there is an order among all possible models of the knowledge base and the revision is made through those lenses.

In parallel, cognitive studies on human BR indicate that people first seek to understand the nature of the inconsistency before they revise their beliefs~\cite{thagard1989explanatory,johnson2004reasoning}. When encountering conflicting information, individuals construct explanations to reconcile these inconsistencies because explanations offer clearer guidance for future actions than mere belief adjustments~\cite{craiknature,keil2006explanation} and play a vital role in communicating one's understanding of the world~\cite{chi1994eliciting,lombrozo2007simplicity}. Importantly, as the literature shows~\cite{johnson2004reasoning,khemlani2013cognitive}, this \emph{explanation-based} approach often leads to BRs that do not always coincide with classical belief base revision~\cite{hansson1994kernel} operators. 

This seeming disconnect between BR frameworks and human cognition creates a gap that is particularly important in the context of human-AI collaboration.~As AI systems play an increasingly important role alongside human users in decision-making tasks, the AI systems need ways to understand and revise their models of the users. These models, which encode the AI system's understanding of what the user believes about a task, are crucial for effective and fluid interactions~\cite{kambhampati2020challenges,sreedharan2021foundations,vasileiou_DR-arg}.  

Our goal in this paper is to empirically investigate how people actually revise inconsistent beliefs and what implications this has for AI systems designed to model human reasoning or interact with humans. Through three comprehensive user studies across different types of scenarios, we show patterns of revision that indicate that people tend to undertake revisions guided by explanations. Our first experiment investigates how people naturally 
perform revisions when encountering inconsistencies, revealing a strong tendency to generate explanations to incorporate the epistemic input in their belief systems. Our second and third experiments examine how people revise their beliefs when \emph{presented with} explanations, showing an even stronger preference for what seem to be non-minimal revisions, in the sense that they choose to discard more information than what would be needed to restore consistency. These findings, which we show are consistent across different types of logical inconsistency, whether the explanations are provided to the users or not, as well as the granularity of the information, call for more research to be focused in modeling human BR, in terms of both theory, e.g., via analyses of existing and novel postulates and operators, and experiments, e.g., via further user studies.

\begin{figure*}[!th]
    \centering
    \includegraphics[width=1\linewidth]{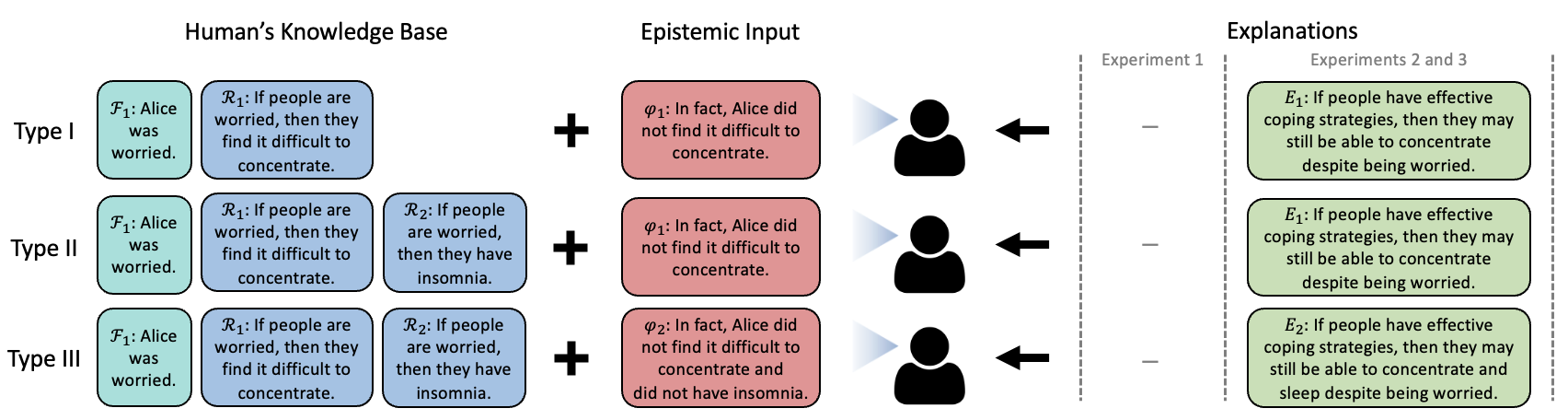}
    \caption{Experimental design 
    for the three problem types, with examples of the human's knowledge base, comprising facts (turquoise boxes) and rules (blue boxes), and the epistemic input (red boxes), along with the explanations (green boxes) provided in Experiments 2 and 3.}
    \label{fig:types}
\end{figure*}

\section{User Studies on Belief Revision in Humans}
\label{sec:empirical}

Our user studies examine how people reason when presented with information that contradicts their beliefs, resolving these inconsistencies into a new set of consistent beliefs.\footnote{Ethics approval was obtained 
from our institution.} In particular, we conducted three experiments: 
\squishlist    %
    \item 
    Experiment 1 explores how people generate explanations when encountering inconsistencies, asking participants to explain why the new information conflicts with their existing beliefs; 
    \item Experiment 2 examines how people revise their beliefs when provided with explanations (those generated in Experiment 1), testing whether explanation-based revision patterns persist when explanations are given rather than self-generated; and 
    \item Experiment 3 investigates whether these revision patterns hold when beliefs are instantiated as specific cases rather than general rules, using grounded versions of the same scenarios with multiple concrete instances.
\squishend

\xhdr{Design Overview:}To carry out our investigation, we selected three types of problems that have been well-studied in the cognitive science literature \cite{elio1997belief,politzer2001belief,byrne2002contradictions}. These problems, illustrated in Figure \ref{fig:types}, present increasingly complex scenarios where new information (the epistemic input) conflicts with existing beliefs (the human's knowledge base), allowing a systematic study on how people handle such inconsistencies. Additional details for all experiments can be found in the supplement.

The first type (\textbf{Type I}) presents participants with a simple scenario containing a conditional (generalization) statement $\fml{R}_1$ and a (categorical) fact $\fml{F}_1$ about a specific case, which together compose the human's knowledge base, and the epistemic input $\varphi_1$ that conflicts with what the knowledge base implies. 

The second type (\textbf{Type II}) increases complexity by introducing an additional conditional statement $\fml{R}_2$, creating a scenario where the epistemic input $\varphi_1$ conflicts  with the consequences of one of the conditional statements and the fact. 

Finally, the third type (\textbf{Type III}) of inconsistency presents the most complex case, where the more complex epistemic input $\varphi_2$ conflicts with the consequences of both conditional statements and the fact. 


In the real world, almost all conditional generalizations about events are susceptible to what psychologists refer to as \textit{disabling conditions} -- conditions describing how the conditional fails~\cite{dieussaert2000initial,elio1997belief,politzer2001belief}. For instance, for the human's knowledge base and epistemic input for Type I in Figure \ref{fig:types}, one might question ``Is it really the case that all people find it difficult to concentrate when they are worried?'' One can easily think of a disabling condition for this conditional, for example, ``people with effective coping strategies may still be able to concentrate despite being worried'', as in the explanation $E_1$ in Figure \ref{fig:types}.  Such explanations, by modifying conditional rules rather than facts, generally result in non-minimal changes to belief systems, as they affect inferences beyond the specific inconsistency at hand.

Due to humans' propensity to envisage disabling conditions, their explanations are more likely to invoke such conditions than to imply that a categorical statement (i.e., the fact) is wrong. Note, however, that this behavior means that they also remove the support for other consequences apart from the one giving rise to the inconsistency. For example, rejecting $\fml{R}_1$ implies rejecting all of its groundings, which means you cannot infer that people find it difficult to concentrate if they are worried, for any instantiation of this rule. 
Note also that, in such cases, both rejecting $\fml{R}_1$ or rejecting $\fml{F}_1$ are equally valid revisions for classical belief base revision~\cite{hansson1994kernel}.

\subsection{Experiment 1}


We recruited 62 participants from the online crowdsourcing platform Prolific \cite{palan2018prolific} across diverse demographics, with the only filter being that they are fluent in English. The participants carried out three different problems of each of the three types (Type I, Type II, and Type III), for a total of nine problems. The statements were taken from common, everyday events including subjects such as economics, intuitive physics, and psychology. The conditional statements in all problems were selected to be highly plausible and interpretable, similar to the high-plausibility category used by Politzer and Carles~\shortcite{politzer2001belief}.

The participants' main task was to explain the inconsistencies presented to them, and we examined the revisions implied by their explanations. After providing their explanations for every problem, each participant was asked how they approached explaining what was presented to them and if they followed any strategies when doing so.

\xhdr{Results:}All participants easily came up with reasons to explain the inconsistencies they encountered. To analyze the results, we employed a coding scheme similar to that by Byrne~\shortcite{byrne2002contradictions}. Following the literature, explanations provided by the participants were categorized into two main categories: (1)~ non-minimal revisions that implied changing or discarding conditional rules, and (2)~ minimal revisions that implied changing or discarding categorical facts). 
Note that these two categories did not involve the epistemic input, which was intended to be accepted by participants in our experiments by design.
The notion of minimality that we are implicitly assuming here is related to the fact that conditional rules contain more information than mere specific facts, and therefore, a revision that seeks to retain as much information as possible would prefer to change or remove facts instead of conditionals. Note that this is not the classical notion of minimality that we find in classical belief base revision. As we mentioned previously, a framework such as Kernel Revision~\cite{hansson1994kernel,ferme2018belief} would consider rejecting any of them equally valid.

Explanations implying revisions of category (1) were either disabling conditions that would prevent the consequences of the generalization, or of the form ``It is not the case that if X then Y'', ``X is not sufficient for Y'', and other similar forms. Explanations that implied categorical statements were rejection to the statements and were of the form ``not X'', ``perhaps not X'', and so on.  This coding scheme classified $89\%$ of the responses. The remaining responses either affirmed or denied the new information, i.e., the epistemic input, or were too vague to classify.

Table~\ref{tab:combined-results} displays the distribution of explanations implying either non-minimal (1) or minimal (2) revisions. The data reveal a compelling trend: a significant majority of explanations across all questions leaned towards revisions that imply removing or changing conditional rules. A Wilcoxon test performed on the aggregated data yielded a p-value significantly smaller than 0.05 ($p \approx 2.96 \times 10^{-50}$), providing robust evidence that the observed proportions are far from what would be expected by random chance. Moreover, effect size measurements (Cohen's $d$) were conducted to quantify the magnitude of these differences, where it was consistently high across all instances.

These results demonstrate that when faced with inconsistencies, participants predominantly created explanations that yield them to discard or modify conditional rules over categorical facts. This suggests that individuals engage deeply in resolving inconsistencies, often opting for more comprehensive explanatory frameworks that modify their existing beliefs to a greater extent than simply choosing an arbitrary minimal set of beliefs to remove. 

\begin{table}[t]
\setlength{\tabcolsep}{4pt}
\centering
\resizebox{1\columnwidth}{!}{
\begin{tabular}{lllllll}
\hline
 & \textbf{Problem Type} & \textbf{Non-minimal} & \textbf{Minimal} & \textbf{Wilcoxon Test} \\
 &  & \textbf{Revision} & \textbf{Revision} & \textbf{(p-value)}  \\
\hline
 & \textit{Type I} & 132 (81.99\%) & 29 (18.01\%) & 4.76 $\times$ 10$^{-16}$ \\
\textbf{Exp. 1} & \textit{Type II} & 140 (86.96\%) & 21 (13.04\%) & 6.69 $\times$ 10$^{-21}$ \\
 & \textit{Type III} & 144 (81.36\%) & 33 (18.64\%) & 7.23 $\times$ 10$^{-17}$ \\
 & Aggregate & 416 (83.37\%) & 83 (16.63\%) & 2.96 $\times$ 10$^{-50}$  \\
\hline
 & \textit{Type I} & 131 (79.39\%) & 34 (20.61\%) & 4.30 $\times$ 10$^{-14}$  \\
\textbf{Exp. 2} & \textit{Type II} & 148 (91.36\%) & 14 (8.64\%) & 6.42 $\times$ 10$^{-26}$  \\
 & \textit{Type III} & 134 (85.90\%) & 22 (14.10\%) & 3.04 $\times$ 10$^{-19}$ \\
 & Aggregate & 413 (85.51\%) & 70 (14.49\%) & 6.52 $\times$ 10$^{-55}$ \\
\hline
 & \textit{Type I} & 113 (71.97\%) & 44 (28.03\%) & 2.68 $\times$ 10$^{-24}$ \\
\textbf{Exp. 3} & \textit{Type II} & 100 (65.36\%) & 53 (34.64\%) & 1.50 $\times$ 10$^{-25}$ \\
 & \textit{Type III} & 104 (64.20\%) & 58 (35.80\%) & 4.61 $\times$ 10$^{-16}$ \\
 & Aggregate & 317 (67.16\%) & 155 (32.84\%) & 1.44 $\times$ 10$^{-52}$ \\
\hline
\end{tabular}
}
\caption{Results from all three \emph{exp}eriments, with \textit{Aggregate} representing combined data from all problem types.
}
\label{tab:combined-results}
\end{table}

\subsection{Experiment 2}

We recruited $60$ participants from the Prolific platform with the same requirements as in Experiment 1. In this study, rather than having the participants generate their own explanations, they were presented with some of the most plausible explanations (that are disabling conditions) created by participants in Experiment 1, and then asked to describe how they would revise their information in light of the explanation. To ensure that they did not discard the explanations, we added some validity to the explanation by telling the participants that the explanation comes from a trustworthy source.

\xhdr{Results:}
We employed a specific coding scheme to analyze how participants chose to revise their beliefs. In accordance with this scheme, participants indicated whether they would \textit{keep}, \textit{discard}, or \textit{alter} the beliefs. We adopted a measure of belief change similar to those used in previous studies \cite{elio1997belief,harman1986change,walsh2009changing,khemlani2013cognitive}, which count the number of beliefs that change their values. 
When choosing to alter a belief, participants were asked to provide details about how they would go about it. As before, we have two categories of revision: non-minimal (category (1)) revisions that discard or alter either a conditional or a combination of more than two statements; and minimal (category (2)) revisions 
that discard or alter the categorical facts. This coding scheme classified $89\%$ of the responses, while the remaining responses either yielded inconsistent revisions (e.g., not revising anything) or were too vague to be classified.

Table \ref{tab:combined-results} provides an overview of the results. The data reveal a clear trend: a significant majority of revisions were non-minimal across all problem types. Participants showed a strong preference for modifying or discarding conditional rules rather than categorical facts, with this pattern being particularly pronounced in Type II problems where over 90\% of participants chose non-minimal revisions. The aggregate analysis across all 413 valid responses showed that 85.51\% were non-minimal revisions. A Wilcoxon test performed on the aggregated data yielded a p-value of $p \approx 6.52 \times 10^-55$, providing strong statistical evidence that this preference was not due to chance.

These findings corroborate those of Experiment 1 and provide evidence that people predominantly opt for non-minimal revisions when presented with explanations. Even when given explanations from a trustworthy source, participants maintained their tendency to make broader changes to their belief systems, suggesting that this preference for explanation-based revision may be a fundamental aspect of how people process inconsistencies.

\subsection{Experiment 3}

We recruited 60 participants from Prolific using the same requirements as in the previous experiments. Building upon the three types of inconsistency problems, we systematically grounded each scenario by creating specific instantiations of the general rules. For Type I scenarios (containing one generalization and one fact, we created four ground conditional statements, and added an additional categorical fact. For Types II and III scenarios (containing two generalizations and one fact), we created eight ground conditional statements (four for each rule), as well as added an additional fact for a total of two.

\xhdr{Results:}
Similarly to Experiment 2, we asked the participants to indicate whether they would keep, discard, or alter the beliefs in light of the explanation. However, given the instantiated nature of the beliefs, we established the following criteria for measuring minimal versus non-minimal revisions. For 
Type I and II 
scenarios, consistency could be restored with a single revision—either modifying one conditional rule or one fact. Therefore, any revision involving more than one belief change was considered non-minimal. For 
Type III 
scenarios, where the inconsistency affected two conditional statements, consistency required at most two revisions. Here, changes to more than two statements were considered non-minimal. To ensure consistent analysis, we counted both discarded and altered statements as changes in our measurement of BR. Using this coding scheme, we classified 87\% of the responses, while the remaining responses yielded inconsistent revisions (e.g., keeping all beliefs).

As can be seen in Table~\ref{tab:combined-results}, our analysis revealed that participants consistently made more extensive changes than the minimum required for consistency. Type I scenarios showed a strong preference for non-minimal revisions, with 71.97\% of participants opting for broader changes ($p \approx 2.68 \times 10^{-24}$). While these scenarios required only one revision for consistency, participants made an average of $2.14$ changes to their beliefs, indicating a clear tendency to revise multiple statements rather than making minimal changes. The gap between minimally required to gain consistency and actual revisions became more pronounced in the subsequent scenarios. In Type II problems, which also required only one revision, 65.36\% of revisions were non-minimal ($p < 10^{-25}$), with participants making an average of $2.87$ changes. Type III scenarios, which required at most two revisions, showed 64.20\% non-minimal revisions ($p < 10^{-16}$), with participants making an average of $3.81$ changes—nearly twice the minimum required. This systematic increase in the average number of changes---from 2.14 in Type I to 3.81 in Type III---suggests that as scenarios become more complex, people make more revisions beyond what is minimally necessary to regain consistency.

\section{Discussion \& Conclusion}
\label{sec:discussion}

For decades researchers in BR have been discussing 
the adequacy of several proposed postulates that can produce revisions that are not aligned with human reasoning~\cite{levi1991fixation,makinson1997force,rott2000two,hansson1999survey,FERME2024109108,HERZIG1999107}.~At the same time, some empirical evidence shows that humans patterns of revision generally do not align with classical approaches to BR~\cite{dutilh2017reasoning,IsbernerK16}.
It is clear, in any case, that if we are to deploy BR mechanisms for modeling humans, the principles that guide them may not accurately capture how humans actually revise their beliefs when faced with inconsistencies, which could have serious consequences for any resulting AI systems which model human reasoning.
 
Our empirical findings suggest that under different circumstances, people may opt for broad revisions to their belief systems, driven by the desire for a more comprehensive understanding and explanation of the information they encounter, and may favor changing or deleting far more information that would seem necessary from a minimality point of view. In particular, our 
experiments revealed a notable propensity among participants to favor what we call explanation-guided revisions when faced with inconsistencies. This preference persisted across different scenarios, suggesting that such an inclination might be a fundamental aspect of human reasoning.

For AI systems aiming to model human reasoning or interact effectively with humans, this means that adhering strictly to minimalism may result in BRs that diverge from human expectations. Instead, such systems should consider incorporating explanation-based approaches to BR that allow for potentially better explanatory understanding. However, implementing such mechanisms in AI systems is not a trivial task. In future work, we plan to develop such an explanation-based BR framework that is inspired by how people revise their beliefs. Addressing these challenges will require interdisciplinary collaboration between researchers in AI, cognitive science, and human-computer interaction. By bridging the gap between formal, knowledge-driven frameworks and human cognitive processes, we can develop more effective and intuitive AI systems for human-AI collaboration \cite{kambhampati2020challenges}.

To conclude, this paper contributes insights for BR theory by empirically exploring how people revise their beliefs in light of inconsistencies. Our findings highlight the importance of explanations in human BR, and the role that different pieces of information play here. Ultimately, BR is not an isolated process, but an integral component of humans' broader quest for explanatory understanding.

\bibliographystyle{kr}
\bibliography{refs}

\begin{thebibliography}{}

\bibitem[\protect\citeauthoryear{Alchourr{\'o}n, G{\"a}rdenfors, and
  Makinson}{1985}]{alchourron1985logic}
Alchourr{\'o}n, C.~E.; G{\"a}rdenfors, P.; and Makinson, D.
\newblock 1985.
\newblock On the logic of theory change: Partial meet contraction and revision
  functions.
\newblock {\em Journal of Symbolic Logic} 50(2):510--530.

\bibitem[\protect\citeauthoryear{Byrne and
  Walsh}{2002}]{byrne2002contradictions}
Byrne, R.~M., and Walsh, C.~R.
\newblock 2002.
\newblock Contradictions and counterfactuals: Generating belief revisions in
  conditional inference.
\newblock In {\em Proceedings of the Annual Meeting of the Cognitive Science
  Society}.

\bibitem[\protect\citeauthoryear{Chi \bgroup et al\mbox.\egroup
  }{1994}]{chi1994eliciting}
Chi, M.~T.; De~Leeuw, N.; Chiu, M.-H.; and LaVancher, C.
\newblock 1994.
\newblock Eliciting self-explanations improves understanding.
\newblock {\em Cognitive Science} 18(3):439--477.

\bibitem[\protect\citeauthoryear{Craik}{1943}]{craiknature}
Craik, K. J.~W.
\newblock 1943.
\newblock {\em The Nature of Explanation}.

\bibitem[\protect\citeauthoryear{Dieussaert \bgroup et al\mbox.\egroup
  }{2000}]{dieussaert2000initial}
Dieussaert, K.; Schaeken, W.; De~Neys, W.; and d'Ydewalle, G.
\newblock 2000.
\newblock Initial belief state as a predictor of belief revision.
\newblock {\em Current Psychology of Cognition}.

\bibitem[\protect\citeauthoryear{Dutilh~Novaes and
  Veluwenkamp}{2017}]{dutilh2017reasoning}
Dutilh~Novaes, C., and Veluwenkamp, H.
\newblock 2017.
\newblock Reasoning biases, non-monotonic logics and belief revision.
\newblock {\em Theoria} 83(1):29--52.

\bibitem[\protect\citeauthoryear{Elio and Pelletier}{1997}]{elio1997belief}
Elio, R., and Pelletier, F.~J.
\newblock 1997.
\newblock Belief change as propositional update.
\newblock {\em Cognitive Science} 21(4):419--460.

\bibitem[\protect\citeauthoryear{Ferm{\'e} and Hansson}{2018}]{ferme2018belief}
Ferm{\'e}, E., and Hansson, S.~O.
\newblock 2018.
\newblock {\em Belief Change: Introduction and Overview}.
\newblock Springer.

\bibitem[\protect\citeauthoryear{Fermé \bgroup et al\mbox.\egroup
  }{2024}]{FERME2024109108}
Fermé, E.; Garapa, M.; Nayak, A.; and Reis, M.~D.
\newblock 2024.
\newblock Relevance, recovery and recuperation: A prelude to ring withdrawal.
\newblock {\em International Journal of Approximate Reasoning} 166:109108.

\bibitem[\protect\citeauthoryear{Hansson}{1994}]{hansson1994kernel}
Hansson, S.~O.
\newblock 1994.
\newblock Kernel contraction.
\newblock {\em Journal of Symbolic Logic} 59(3):845--859.

\bibitem[\protect\citeauthoryear{Hansson}{1999}]{hansson1999survey}
Hansson, S.~O.
\newblock 1999.
\newblock A survey of non-prioritized belief revision.
\newblock {\em Erkenntnis} 50(2-3):413--427.

\bibitem[\protect\citeauthoryear{Harman}{1986}]{harman1986change}
Harman, G.
\newblock 1986.
\newblock {\em Change in View: Principles of Reasoning}.
\newblock The MIT Press.

\bibitem[\protect\citeauthoryear{Herzig and Rifi}{1999}]{HERZIG1999107}
Herzig, A., and Rifi, O.
\newblock 1999.
\newblock Propositional belief base update and minimal change.
\newblock {\em Artificial Intelligence} 115(1):107--138.

\bibitem[\protect\citeauthoryear{Isberner and
  Kern{-}Isberner}{2016}]{IsbernerK16}
Isberner, M., and Kern{-}Isberner, G.
\newblock 2016.
\newblock A formal model of plausibility monitoring in language comprehension.
\newblock In Markov, Z., and Russell, I., eds., {\em Proceedings of the
  Twenty-Ninth International Florida Artificial Intelligence Research Society
  Conference, {FLAIRS} 2016, Key Largo, Florida, USA, May 16-18, 2016},
  662--667.
\newblock {AAAI} Press.

\bibitem[\protect\citeauthoryear{Johnson-Laird, Girotto, and
  Legrenzi}{2004}]{johnson2004reasoning}
Johnson-Laird, P.~N.; Girotto, V.; and Legrenzi, P.
\newblock 2004.
\newblock Reasoning from inconsistency to consistency.
\newblock {\em Psychological Review} 111(3):640.

\bibitem[\protect\citeauthoryear{Kambhampati}{2020}]{kambhampati2020challenges}
Kambhampati, S.
\newblock 2020.
\newblock Challenges of human-aware {AI} systems: {AAAI} presidential address.
\newblock {\em AI Magazine} 41(3):3--17.

\bibitem[\protect\citeauthoryear{Keil}{2006}]{keil2006explanation}
Keil, F.~C.
\newblock 2006.
\newblock Explanation and understanding.
\newblock {\em Annual Review of Psychology} 57:227--254.

\bibitem[\protect\citeauthoryear{Khemlani and
  Johnson-Laird}{2013}]{khemlani2013cognitive}
Khemlani, S., and Johnson-Laird, P.
\newblock 2013.
\newblock Cognitive changes from explanations.
\newblock {\em Journal of Cognitive Psychology} 25(2):139--146.

\bibitem[\protect\citeauthoryear{Levi}{1991}]{levi1991fixation}
Levi, I.
\newblock 1991.
\newblock {\em The fixation of belief and its undoing: Changing beliefs through
  inquiry}.
\newblock Cambridge University Press.

\bibitem[\protect\citeauthoryear{Lombrozo}{2007}]{lombrozo2007simplicity}
Lombrozo, T.
\newblock 2007.
\newblock Simplicity and probability in causal explanation.
\newblock {\em Cognitive Psychology} 55(3):232--257.

\bibitem[\protect\citeauthoryear{Makinson}{1997}]{makinson1997force}
Makinson, D.
\newblock 1997.
\newblock {\em On the force of some apparent counterexamples to recovery}.
\newblock na.

\bibitem[\protect\citeauthoryear{Palan and Schitter}{2018}]{palan2018prolific}
Palan, S., and Schitter, C.
\newblock 2018.
\newblock Prolific: A subject pool for online experiments.
\newblock {\em Journal of Behavioral and Experimental Finance} 17:22--27.

\bibitem[\protect\citeauthoryear{Politzer and
  Carles}{2001}]{politzer2001belief}
Politzer, G., and Carles, L.
\newblock 2001.
\newblock Belief revision and uncertain reasoning.
\newblock {\em Thinking \& Reasoning} 7(3):217--234.

\bibitem[\protect\citeauthoryear{Rott}{2000}]{rott2000two}
Rott, H.
\newblock 2000.
\newblock Two dogmas of belief revision.
\newblock {\em Journal of Philosophy} 97(9):503--522.

\bibitem[\protect\citeauthoryear{Sreedharan, Chakraborti, and
  Kambhampati}{2021}]{sreedharan2021foundations}
Sreedharan, S.; Chakraborti, T.; and Kambhampati, S.
\newblock 2021.
\newblock Foundations of explanations as model reconciliation.
\newblock {\em Artificial Intelligence} 301:103558.

\bibitem[\protect\citeauthoryear{Thagard}{1989}]{thagard1989explanatory}
Thagard, P.
\newblock 1989.
\newblock Explanatory coherence.
\newblock {\em Behavioral and Brain Sciences} 12(3):435--467.

\bibitem[\protect\citeauthoryear{Vasileiou \bgroup et al\mbox.\egroup
  }{2024}]{vasileiou_DR-arg}
Vasileiou, S.~L.; Kumar, A.; Yeoh, W.; Son, T.~C.; and Toni, F.
\newblock 2024.
\newblock Dialectical reconciliation via structured argumentative dialogues.
\newblock In {\em Proceedings of the International Conference on Principles of
  Knowledge Representation and Reasoning (KR)},  777--787.

\bibitem[\protect\citeauthoryear{Walsh and
  Johnson-Laird}{2009}]{walsh2009changing}
Walsh, C.~R., and Johnson-Laird, P.
\newblock 2009.
\newblock Changing your mind.
\newblock {\em Memory \& Cognition} 37(5):624--631.

\end{thebibliography}

\newpage
\clearpage

\appendix

\section{User Study Experiments}

In what follows, we describe the three experiments that we undertook, and provide some additional analysis of the results.

\subsection{Experiment 1}




Participants in this experiment were given the following information and instructions: 
\begin{quote}
\textit{You will be presented with a series of everyday common scenarios. In each scenario, you will be presented with information from two or three different speakers talking about some specific things. You will then be given some additional information that you know, for a fact, to be true. Your task, in essence, is to explain what is going on.}
\end{quote}

\begin{itemize}
    \item \textit{\textbf{Read carefully:} For each scenario, read all the information very carefully.}
    \item \textit{\textbf{Explain}: Think about how to explain the fact. In other words, ask yourself: why does the fact conflict with the information provided by the speakers? Answer in your own words.}
    \item \textit{\textbf{No Right or Wrong Answers}: This study aims to understand your personal thought process. There are no right or wrong answers. Choose what feels most accurate to you.}
    \item \textit{\textbf{Pace Yourself}: While there's no strict time limit, try to spend a reasonable amount of time on each scenario—neither rushing through nor overthinking too much.}

\end{itemize}

Afterwards, the participants saw the following nine scenarios, and their task was to answer the corresponding question:\\

\noindent \textbf{Scenario 1 (Type I):}
\begin{itemize}
    \item \textbf{$\fml{R}_1$}: \textit{If a drink contains sugar, then it gives you energy.}
    \item \textbf{$\fml{F}_1$}: \textit{This drink contains sugar.}
    \item \textbf{Fact}: \textit{In fact, it doesn't give you energy.}
\end{itemize}

\textit{Why does the drink not give you energy?}

\medskip \noindent \textbf{Scenario 2 (Type I):}
\begin{itemize}
    \item \textbf{$\fml{R}_1$}: \textit{If sales go up, then profits improve.}
    \item \textbf{$\fml{F}_1$}: \textit{The sales went up.}
    \item \textbf{Fact}: \textit{In fact, the profits did not go up.}
\end{itemize}

\textit{Why did the sales not go up?}

\medskip \noindent \textbf{Scenario 3 (Type I):}
\begin{itemize}
    \item \textbf{$\fml{R}_1$}: \textit{If people have a fever, then they have a high temperature.}
    \item \textbf{$\fml{F}_1$}: \textit{Maria had a fever.}
    \item \textbf{Fact}: \textit{In fact, Maria did not have a high temperature.}
\end{itemize}

\textit{Why did Maria not have a high temperature?}

\medskip \noindent \textbf{Scenario 4 (Type II):}
\begin{itemize}
    \item \textbf{$\fml{R}_1$}: \textit{If there is very loud music, then it is difficult to have a conversation.}
    \item \textbf{$\fml{R}_2$}: \textit{If there is very loud music, then the neighbors complain.}
    \item  \textbf{$\fml{F}_1$}: \textit{The music was loud.}
    \item \textbf{Fact}: \textit{In fact, the neighbors did not complain.}
\end{itemize}

\textit{Why did the neighbors not complain?}

\medskip \noindent \textbf{Scenario 5 (Type II):}
\begin{itemize}
    \item \textbf{$\fml{R}_1$}: \textit{If people are worried, then they find it difficult to concentrate.}
    \item \textbf{$\fml{R}_2$}: \textit{If people are worried, then they have insomnia.}
    \item  \textbf{$\fml{F}_1$}: \textit{Alice was worried.}
    \item \textbf{Fact}: \textit{In fact, Alice did not find it difficult to concentrate.}
\end{itemize}

\textit{Why did Alice not find it difficult to concentrate?}

\medskip \noindent \textbf{Scenario 6 (Type II):}
\begin{itemize}
    \item \textbf{$\fml{R}_1$}: \textit{If you follow this diet, then you lose weight.}
    \item \textbf{$\fml{R}_2$}: \textit{If you follow this diet, then you have a good supply of iron}
    \item  \textbf{$\fml{F}_1$}: \textit{John followed this diet.}
    \item \textbf{Fact}: \textit{In fact, John did not lose weight.}
\end{itemize}

\textit{Why did John not lose weight?}

\medskip \noindent \textbf{Scenario 7 (Type III):}
\begin{itemize}
    \item \textbf{$\fml{R}_1$}: \textit{If someone is very kind to you, then you like that person.}
    \item \textbf{$\fml{R}_2$}: \textit{If someone is very kind to you, then you are kind in return.}
    \item  \textbf{$\fml{F}_1$}: \textit{Jocko is very kind to Kristen.}
    \item \textbf{Fact}: \textit{In fact, Kristen did not like Jocko, and she were not kind in return.}
\end{itemize}

\textit{Why did Kristen not like Jocko and was not kind to him?}

\medskip \noindent \textbf{Scenario 8 (Type III):}
\begin{itemize}
    \item \textbf{$\fml{R}_1$}: \textit{If a match is struck, then it produces light.}
    \item \textbf{$\fml{R}_2$}: \textit{If a match is struck, then it gives off smoke.}
    \item  \textbf{$\fml{F}_1$}: \textit{Mary struck a match.}
    \item \textbf{Fact}: \textit{In fact, the match produced no light, and it did not give off smoke.}
\end{itemize}

\textit{Why did the match produce no light and gave off no smoke?}

\medskip \noindent \textbf{Scenario 9 (Type III):}
\begin{itemize}
    \item \textbf{$\fml{R}_1$}: \textit{If people are nervous, then their hands shake.}
    \item \textbf{$\fml{R}_2$}: \textit{ If people are nervous, then they get butterflies in their stomach.}
    \item  \textbf{$\fml{F}_1$}: \textit{Patrick was nervous.}
    \item \textbf{Fact}: \textit{In fact, Patrick's hands did not shake, and he didn't get butterflies in his stomach.}
\end{itemize}

\textit{Why did Patrick's hands not shake and he didn't get butterflies in his stomach?}

\begin{figure*}[!t]
    \centering
    \includegraphics[width=0.9\textwidth]{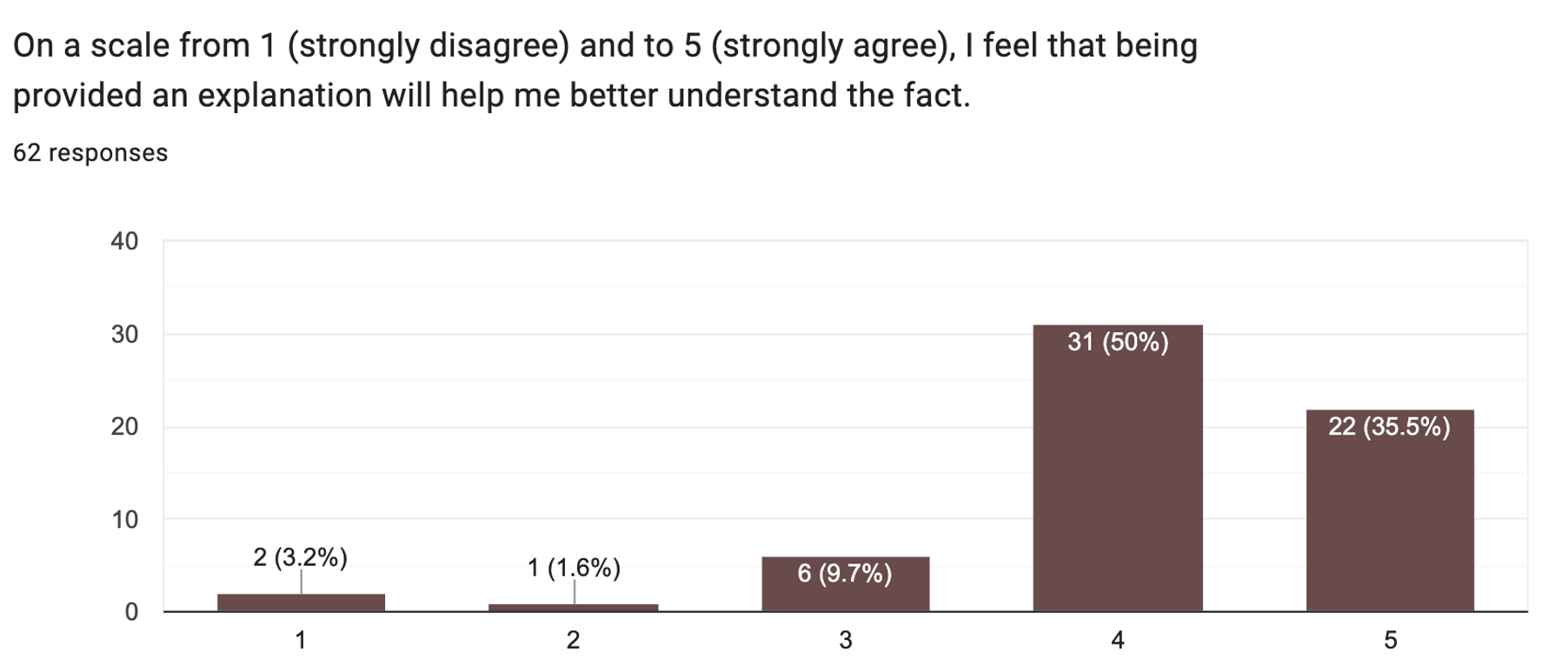}
    \caption{Distribution of responses to the Likert-type question Q2 in Experiment 1.}
    \label{fig:exp1:likert}
\end{figure*}

\medskip
After going through all nine scenarios, the participants were finally asked the following two questions:

\smallskip \noindent \textbf{Q1:} \textit{Describe in your own words how you approached explaining what was going on. Was there a specific reason why you chose to retain or discard certain information?} 

\smallskip \noindent \textbf{Q2:} \textit{On a scale from 1 (strongly disagree) and to 5 (strongly agree), I feel that being provided an explanation will help me better understand the fact.}

\smallskip \noindent Figure~\ref{fig:exp1:likert} shows the distribution of the Likert question (Q2).

\subsection{Experiment 2}

In this experiment, the participants saw the same nine scenarios as in Experiment 1 but with a corresponding explanation that explains the inconsistency. Their task was to describe how they would revise their information in light of the given explanation. To ensure that they will not discard the explanation, they were informed that the explanation is trustworthy.


The scenarios the participants saw can be seen below:

\noindent \textbf{Scenario 1 (Type I):}
\begin{itemize}
    \item \textbf{$\fml{R}_1$}: \textit{If a drink contains sugar, then it gives you energy.}
    \item \textbf{$\fml{F}_1$}: \textit{This drink contains sugar.}
    \item \textbf{Fact}: \textit{In fact, it doesn't give you energy.}
    \item \textbf{Explanation:} \textit{If a person has metabolic disorders, then a sugary drink may not provide energy.}

\end{itemize}

\textit{Why does the drink not give you energy?}

\medskip \noindent \textbf{Scenario 2 (Type I):}
\begin{itemize}
    \item \textbf{$\fml{R}_1$}: \textit{If sales go up, then profits improve.}
    \item \textbf{$\fml{F}_1$}: \textit{The sales went up.}
    \item \textbf{Fact}: \textit{In fact, the profits did not go up.}
    \item \textbf{Explanation}: \textit{If expenses rise at a faster rate than sales, then an increase in sales may not lead to improved profits.}

\end{itemize}

\textit{Why did the sales not go up?}

\medskip \noindent \textbf{Scenario 3 (Type I):}
\begin{itemize}
    \item \textbf{$\fml{R}_1$}: \textit{If people have a fever, then they have a high temperature.}
    \item \textbf{$\fml{F}_1$}: \textit{Maria had a fever.}
    \item \textbf{Fact}: \textit{In fact, Maria did not have a high temperature.}
    \item \textbf{Explanation}: \textit{If a person has taken antipyretics, then they may not have a high temperature.}
\end{itemize}

\medskip \noindent \textbf{Scenario 4 (Type II):}
\begin{itemize}
    \item \textbf{$\fml{R}_1$}: \textit{If there is very loud music, then it is difficult to have a conversation.}
    \item \textbf{$\fml{R}_2$}: \textit{If there is very loud music, then the neighbors complain.}
    \item  \textbf{$\fml{F}_1$}: \textit{The music was loud.}
    \item \textbf{Fact}: \textit{In fact, the neighbors did not complain.}
    \item \textbf{Explanation:} \textit{If the neighbors are away on vacations, then very loud music does not lead to complaints.}
\end{itemize}

\medskip \noindent \textbf{Scenario 5 (Type II):}
\begin{itemize}
    \item \textbf{$\fml{R}_1$}: \textit{If people are worried, then they find it difficult to concentrate.}
    \item \textbf{$\fml{R}_2$}: \textit{If people are worried, then they have insomnia.}
    \item  \textbf{$\fml{F}_1$}: \textit{Alice was worried.}
    \item \textbf{Fact}: \textit{In fact, Alice did not find it difficult to concentrate.}

    \item  \textbf{Explanation:} \textit{If people have effective coping strategies, then they may still be able to concentrate despite being worried.}
\end{itemize}

\medskip \noindent \textbf{Scenario 6 (Type II):}
\begin{itemize}
    \item \textbf{$\fml{R}_1$}: \textit{If you follow this diet, then you lose weight.}
    \item \textbf{$\fml{R}_2$}: \textit{If you follow this diet, then you have a good supply of iron}
    \item  \textbf{$\fml{F}_1$}: \textit{John followed this diet.}
    \item \textbf{Fact}: \textit{In fact, John did not lose weight.}

    \item  \textbf{Explanation:} \textit{If people have metabolic imbalances, then following a particular diet may not result in weight loss.}
\end{itemize}

\medskip \noindent \textbf{Scenario 7 (Type III):}
\begin{itemize}
    \item \textbf{$\fml{R}_1$}: \textit{If someone is very kind to you, then you like that person.}
    \item \textbf{$\fml{R}_2$}: \textit{If someone is very kind to you, then you are kind in return.}
    \item  \textbf{$\fml{F}_1$}: \textit{Jocko is very kind to Kristen.}
    \item \textbf{Fact}: \textit{In fact, Kristen did not like Jocko, and she were not kind in return.}

    \item \textbf{Explanation:} \textit{If people have had negative past experiences with someone, then they may not like that person or reciprocate kindness despite the person being kind to them.}
    
\end{itemize}

\medskip \noindent \textbf{Scenario 8 (Type III):}
\begin{itemize}
    \item \textbf{$\fml{R}_1$}: \textit{If a match is struck, then it produces light.}
    \item \textbf{$\fml{R}_2$}: \textit{If a match is struck, then it gives off smoke.}
    \item  \textbf{$\fml{F}_1$}: \textit{Mary struck a match.}
    \item \textbf{Fact}: \textit{In fact, the match produced no light, and it did not give off smoke.}

    \item \textbf{Explanation:} \textit{If the match is wet, then it will neither produce light nor give off smoke.}
\end{itemize}

\medskip \noindent \textbf{Scenario 9 (Type III):}
\begin{itemize}
    \item \textbf{$\fml{R}_1$}: \textit{If people are nervous, then their hands shake.}
    \item \textbf{$\fml{R}_2$}: \textit{ If people are nervous, then they get butterflies in their stomach.}
    \item  \textbf{$\fml{F}_1$}: \textit{Patrick was nervous.}
    \item \textbf{Fact}: \textit{In fact, Patrick's hands did not shake, and he didn't get butterflies in his stomach.}

    \item \textbf{Explanation:} \textit{If individuals have practiced stress-management techniques, then they may not exhibit shaky hands or butterflies in the stomach when nervous.}
\end{itemize}

After each single scenario, the participants answered the following question:

\smallskip \noindent \textit{Describe in your own words how you will revise the information. Was there a specific reason why you chose to retain or discard information from the speakers? To be brief, you can write: keep \textbf{$\fml{R}_1$}, discard \textbf{$\fml{R}_1$}, alter \textbf{$\fml{R}_1$}, and so on (if you alter, please describe how).}

\smallskip 

\subsection{Experiment 3}

\begin{figure*}[!t]
    \centering
    \includegraphics[width=0.9\textwidth]{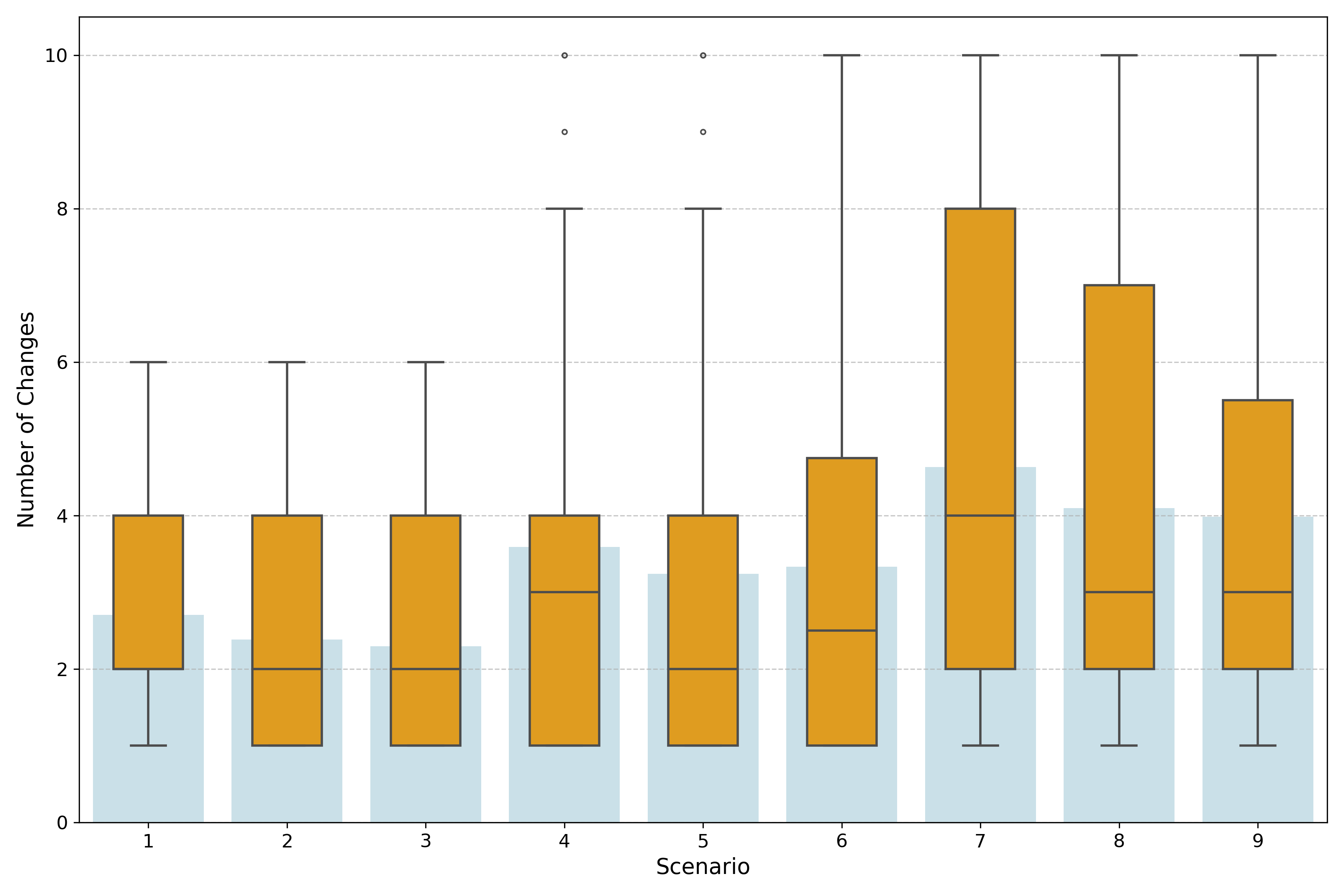}
    \caption{Average number of belief changes per scenario in Experiment 3.}
    \label{fig:exp3:avg-changes}
\end{figure*}

\begin{figure*}[!t]
    \centering
    \includegraphics[width=0.9\textwidth]{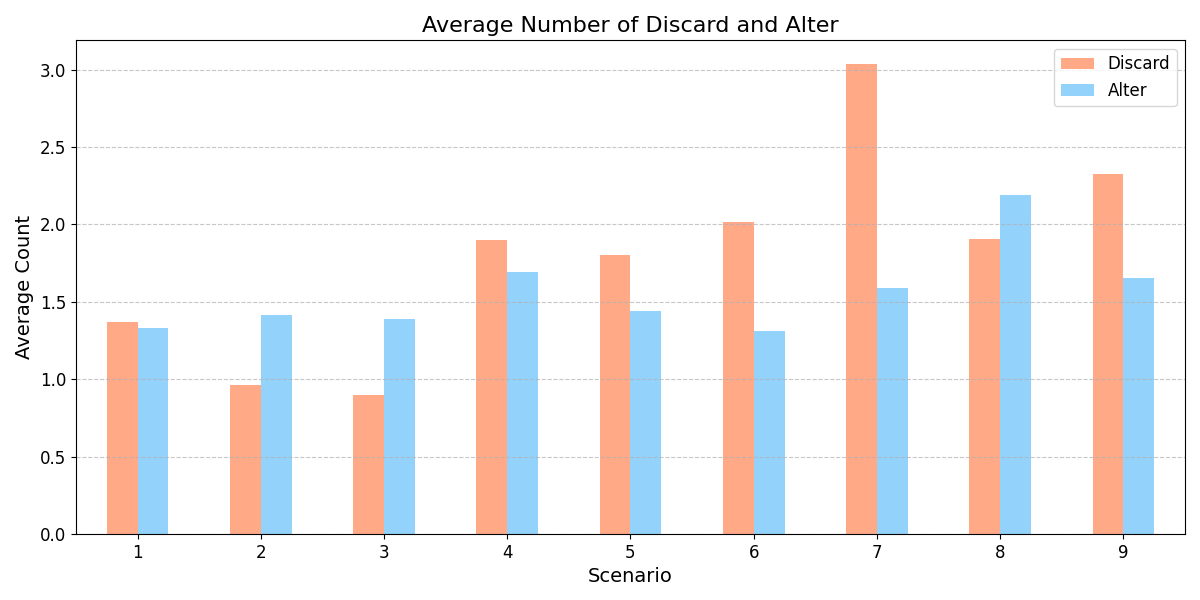}
    \caption{Average number of times statements were discarded and altered in each scenario of Experiment 3.}
    \label{fig:exp3:avg-discard-alter}
\end{figure*}

\begin{figure*}[t!]
    \centering
    \begin{subfigure}[b]{0.32\textwidth}
        \centering
        \includegraphics[width=\textwidth]{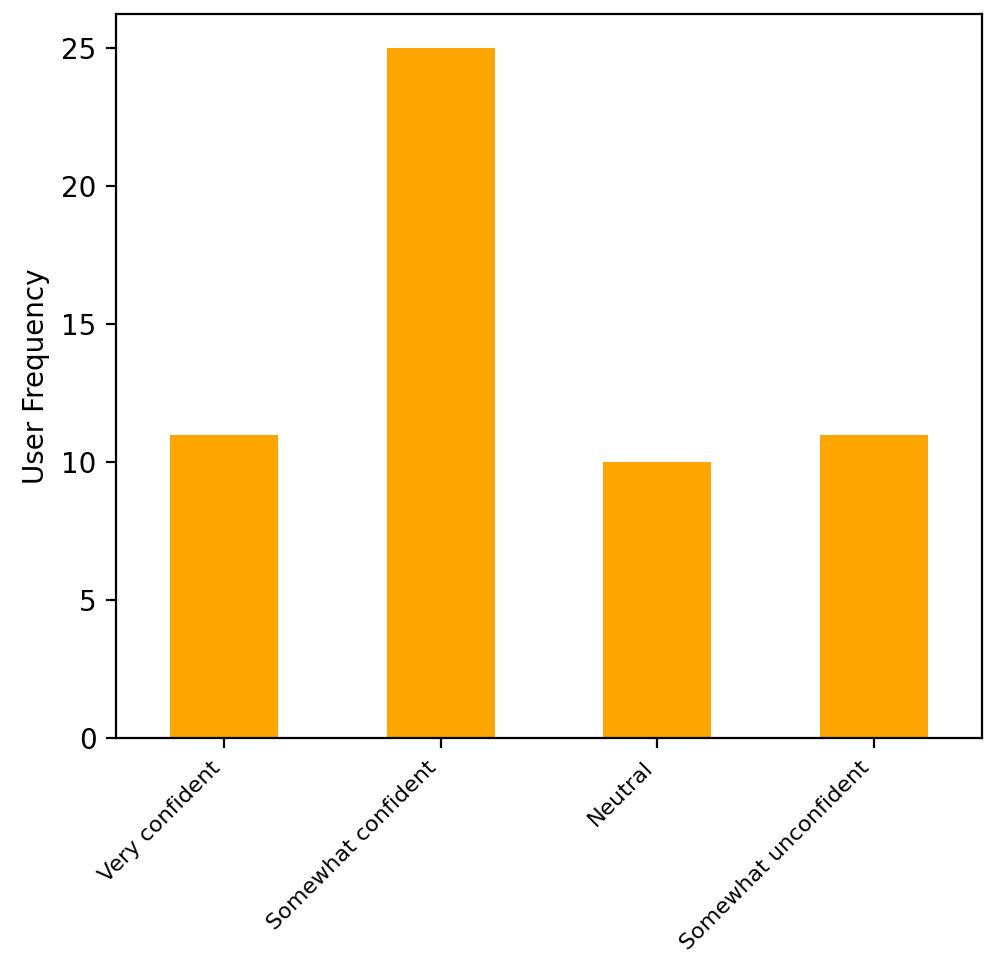}
        \caption{Confidence in revision decisions}
        \label{fig:confidence}
    \end{subfigure}
    \hfill
    \begin{subfigure}[b]{0.32\textwidth}
        \centering
        \includegraphics[width=\textwidth]{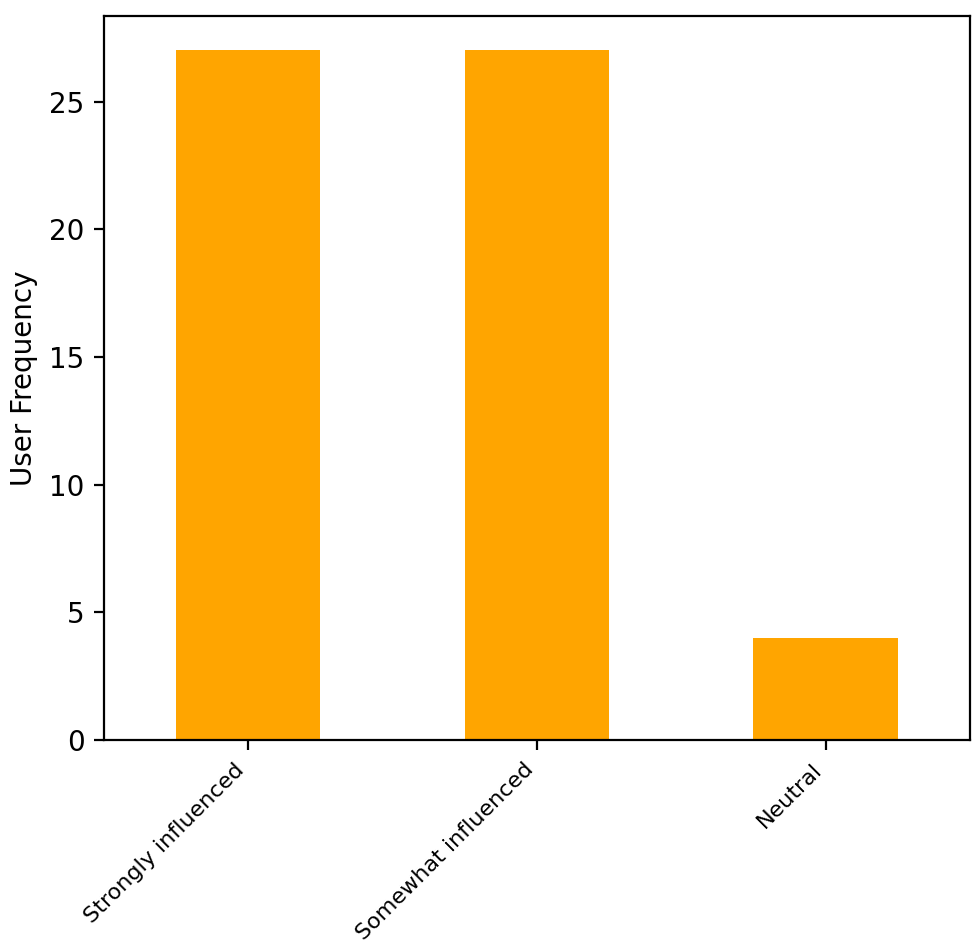}
        \caption{Influence of explanations}
        \label{fig:influence}
    \end{subfigure}
    \hfill
    \begin{subfigure}[b]{0.32\textwidth}
        \centering
        \includegraphics[width=\textwidth]{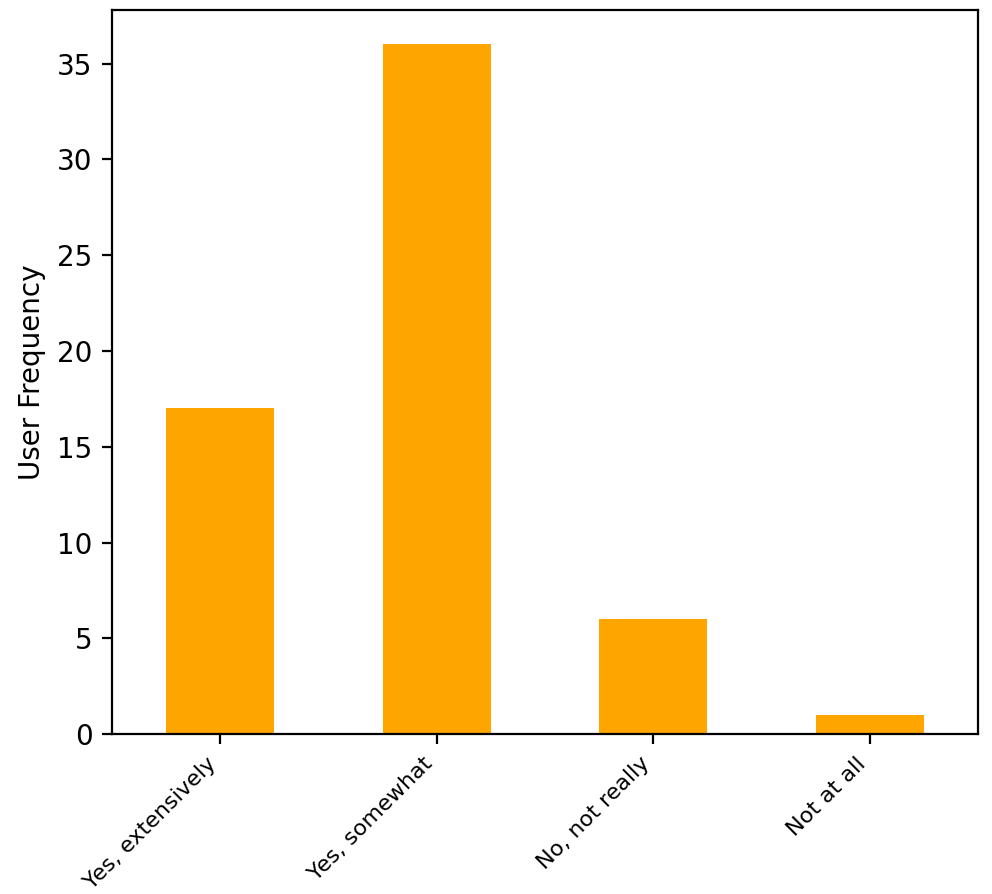}
        \caption{Consideration of broader applications}
        \label{fig:generalization}
    \end{subfigure}
    \caption{Qualitative feedback from participants in Experiment 3 showing (a) their confidence levels in revision decisions, (b) the extent to which explanations influenced their decisions, and (c) whether they considered how explanations might apply beyond specific contradictions.}
    \label{fig:qualitative}
\end{figure*}

In this experiments, the participants saw the nine scenarios depicted below, and their task was to indicate which statements they will discard, alter, or keep. To better understand participants' decision-making processes, we also collected qualitative data at the end of the experiment by asking the participants three subjective questions, such as about their confidence in their revision decisions and if they considered how the explanation might apply to beliefs beyond the specific contradiction. Figure~\ref{fig:exp3:avg-changes} plots the distribution of average number of belief changes in the nine scenarios, while Figure~\ref{fig:exp3:avg-discard-alter} shows the average number of times beliefs were discarded or altered by participants. Finally, Figure~\ref{fig:qualitative} shows the distribution of the answers to three subjective questions.


\medskip \noindent \textbf{Scenario 1 (Type I):}
\begin{itemize}
    \item \textbf{$\fml{R}_1$}: \textit{If orange juice contains sugar, then orange juice gives Tom energy.}
    \item \textbf{$\fml{R}_2$}: \textit{If orange juice contains sugar, then orange juice gives Sarah energy.}
    \item \textbf{$\fml{R}_3$}: \textit{If cola contains sugar, then cola gives Tom energy.}

    \item \textbf{$\fml{R}_4$:} \textit{If cola contains sugar, then cola gives Sarah energy.}
    
    \item  \textbf{$\fml{F}_1$}: \textit{This orange juice contains sugar.}
    \item \textbf{$\fml{F}_2$}: \textit{ This cola contains sugar.}
    
    \item \textbf{Fact}: \textit{In fact, the orange juice did not give Tom energy.}

    \item \textbf{Explanation:} \textit{If a person has metabolic disorders, then a sugary drink may not provide energy.}
\end{itemize}

\medskip \noindent \textbf{Scenario 2 (Type I):}
\begin{itemize}
\item \textbf{$\fml{R}_1$}: \textit{If electronics sales increase, then electronics profits improve for Store A.}
\item \textbf{$\fml{R}_2$}: \textit{If electronics sales increase, then electronics profits improve for Store B.}
\item \textbf{$\fml{R}_3$}: \textit{If clothing sales increase, then clothing profits improve for Store A.}
\item \textbf{$\fml{R}_4$}: \textit{If clothing sales increase, then clothing profits improve for Store B.}
\item \textbf{$\fml{F}_1$}: \textit{Electronics sales went up.}
\item \textbf{$\fml{F}_2$}: \textit{Clothing sales went up.}

\item \textbf{Fact}: \textit{In fact, electronics profits did not improve for Store A.}

\item \textbf{Explanation}: \textit{If expenses rise at a faster rate than sales, then an increase in sales may not lead to improved profits.}
\end{itemize}

\medskip \noindent \textbf{Scenario 3 (Type I):}
\begin{itemize}
\item \textbf{$\fml{R}_1$}: \textit{If morning fever occurs, then morning fever causes Maria's temperature to rise.}
\item \textbf{$\fml{R}_2$}: \textit{If morning fever occurs, then morning fever causes Robert's temperature to rise.}
\item \textbf{$\fml{R}_3$}: \textit{If evening fever occurs, then evening fever causes Maria's temperature to rise.}
\item \textbf{$\fml{R}_4$}: \textit{If evening fever occurs, then evening fever causes Robert's temperature to rise.}
\item \textbf{$\fml{F}_1$}: \textit{Morning fever occurred.}
\item \textbf{$\fml{F}_2$}: \textit{Evening fever occurred.}

\item \textbf{Fact}: \textit{In fact, Maria's temperature did not rise during her morning fever.}

\item \textbf{Explanation}: \textit{If a person has taken antipyretics, then they may not have a high temperature.}
\end{itemize}

\medskip \noindent \textbf{Scenario 4 (Type II):}
\begin{itemize}
\item \textbf{$\fml{R}_1$}: \textit{If rock music is loud, then rock music makes conversation difficult for the Browns.}
\item \textbf{$\fml{R}_2$}: \textit{If rock music is loud, then rock music makes conversation difficult for the Smiths.}
\item \textbf{$\fml{R}_3$}: \textit{If electronic music is loud, then electronic music makes conversation difficult for the Browns.}
\item \textbf{$\fml{R}_4$}: \textit{If electronic music is loud, then electronic music makes conversation difficult for the Smiths.}
\item \textbf{$\fml{R}_5$}: \textit{If rock music is loud, then rock music makes the Browns complain.}
\item \textbf{$\fml{R}_6$}: \textit{If rock music is loud, then rock music makes the Smiths complain.}
\item \textbf{$\fml{R}_7$}: \textit{If electronic music is loud, then electronic music makes the Browns complain.}
\item \textbf{$\fml{R}_8$}: \textit{If electronic music is loud, then electronic music makes the Smiths complain.}
\item \textbf{$\fml{F}_1$}: \textit{The rock music was loud.}
\item \textbf{$\fml{F}_2$}: \textit{The electronic music was loud.}

\item \textbf{Fact}: \textit{In fact, the Browns did not complain about the rock music.}
\item \textbf{Explanation}: \textit{If the neighbors are away on vacations, then very loud music does not lead to complaints.}
\end{itemize}
\medskip \noindent \textbf{Scenario 5 (Type II):}
\begin{itemize}
\item \textbf{$\fml{R}_1$}: \textit{If presentation worry occurs, then presentation worry makes Alice lose concentration.}
\item \textbf{$\fml{R}_2$}: \textit{If presentation worry occurs, then presentation worry makes John lose concentration.}
\item \textbf{$\fml{R}_3$}: \textit{If exam worry occurs, then exam worry makes Alice lose concentration.}
\item \textbf{$\fml{R}_4$}: \textit{If exam worry occurs, then exam worry makes John lose concentration.}
\item \textbf{$\fml{R}_5$}: \textit{If presentation worry occurs, then presentation worry gives Alice insomnia.}
\item \textbf{$\fml{R}_6$}: \textit{If presentation worry occurs, then presentation worry gives John insomnia.}
\item \textbf{$\fml{R}_7$}: \textit{If exam worry occurs, then exam worry gives Alice insomnia.}
\item \textbf{$\fml{R}_8$}: \textit{If exam worry occurs, then exam worry gives John insomnia.}
\item \textbf{$\fml{F}_1$}: \textit{Presentation worry occurred.}
\item \textbf{$\fml{F}_2$}: \textit{Exam worry occurred.}

\item \textbf{Fact}: \textit{In fact, Alice did not lose concentration during her presentation.}
\item \textbf{Explanation}: \textit{If people have effective coping strategies, then they may still be able to concentrate despite being worried.}
\end{itemize}
\medskip \noindent \textbf{Scenario 6 (Type II):}
\begin{itemize}
\item \textbf{$\fml{R}_1$}: \textit{If Mediterranean diet is followed, then Mediterranean diet helps David lose weight.}
\item \textbf{$\fml{R}_2$}: \textit{If Mediterranean diet is followed, then Mediterranean diet helps Emma lose weight.}
\item \textbf{$\fml{R}_3$}: \textit{If Keto diet is followed, then Keto diet helps David lose weight.}
\item \textbf{$\fml{R}_4$}: \textit{If Keto diet is followed, then Keto diet helps Emma lose weight.}
\item \textbf{$\fml{R}_5$}: \textit{If Mediterranean diet is followed, then Mediterranean diet gives David good iron levels.}
\item \textbf{$\fml{R}_6$}: \textit{If Mediterranean diet is followed, then Mediterranean diet gives Emma good iron levels.}
\item \textbf{$\fml{R}_7$}: \textit{If Keto diet is followed, then Keto diet gives David good iron levels.}
\item \textbf{$\fml{R}_8$}: \textit{If Keto diet is followed, then Keto diet gives Emma good iron levels.}
\item \textbf{$\fml{F}_1$}: \textit{Mediterranean diet was followed}
\item \textbf{$\fml{F}_2$}: \textit{Keto diet was followed}

\item \textbf{Fact}: \textit{In fact, David did not lose weight on the Mediterranean diet.}
\item \textbf{Explanation}: \textit{If people have metabolic imbalances, then following a particular diet may not result in weight loss.}
\end{itemize}
\medskip \noindent \textbf{Scenario 7 (Type III):}
\begin{itemize}
\item \textbf{$\fml{R}_1$}: \textit{If classroom kindness occurs, then classroom kindness makes Jocko like Kristen.}
\item \textbf{$\fml{R}_2$}: \textit{If classroom kindness occurs, then classroom kindness makes Kristen like Jocko.}
\item \textbf{$\fml{R}_3$}: \textit{If office kindness occurs, then office kindness makes Jocko like Kristen.}
\item \textbf{$\fml{R}_4$}: \textit{If office kindness occurs, then office kindness makes Kristen like Jocko.}
\item \textbf{$\fml{R}_5$}: \textit{If classroom kindness occurs, then classroom kindness makes Jocko kind in return.}
\item \textbf{$\fml{R}_6$}: \textit{If classroom kindness occurs, then classroom kindness makes Kristen kind in return.}
\item \textbf{$\fml{R}_7$}: \textit{If office kindness occurs, then office kindness makes Jocko kind in return.}
\item \textbf{$\fml{R}_8$}: \textit{If office kindness occurs, then office kindness makes Kristen kind in return.}
\item \textbf{$\fml{F}_1$}: \textit{Classroom kindness occurred.}
\item \textbf{$\fml{F}_2$}: \textit{Office kindness occurred.}

\item \textbf{Fact}: \textit{In fact, Kristen did not like Jocko despite his classroom kindness, and she was not kind in return.}
\item \textbf{Explanation}: \textit{If people have had negative past experiences with someone, then they may not like that person or reciprocate kindness despite the person being kind to them.}
\end{itemize}
\medskip \noindent \textbf{Scenario 8 (Type III):}
\begin{itemize}
\item \textbf{$\fml{R}_1$}: \textit{If wooden match is struck, then wooden match produces light for Jane.}
\item \textbf{$\fml{R}_2$}: \textit{If wooden match is struck, then wooden match produces light for Peter.}
\item \textbf{$\fml{R}_3$}: \textit{If safety match is struck, then safety match produces light for Jane.}
\item \textbf{$\fml{R}_4$}: \textit{If safety match is struck, then safety match produces light for Peter.}
\item \textbf{$\fml{R}_5$}: \textit{If wooden match is struck, then wooden match gives off smoke for Jane.}
\item \textbf{$\fml{R}_6$}: \textit{If wooden match is struck, then wooden match gives off smoke for Peter.}
\item \textbf{$\fml{R}_7$}: \textit{If safety match is struck, then safety match gives off smoke for Jane.}
\item \textbf{$\fml{R}_8$}: \textit{If safety match is struck, then safety match gives off smoke for Peter.}
\item \textbf{$\fml{F}_1$}: \textit{Wooden match was struck.}
\item \textbf{$\fml{F}_2$}: \textit{Safety match was struck.}

\item \textbf{Fact}: \textit{In fact, the wooden match did not produce light or smoke for Jane.}
\item \textbf{Explanation}: \textit{If a match is wet, then it will neither produce light nor give off smoke.}
\end{itemize}
\medskip \noindent \textbf{Scenario 9 (Type III):}
\begin{itemize}
\item \textbf{$\fml{R}_1$}: \textit{If speech nervousness occurs, then speech nervousness makes Patrick's hands shake.}
\item \textbf{$\fml{R}_2$}: \textit{If speech nervousness occurs, then speech nervousness makes Diana's hands shake.}
\item \textbf{$\fml{R}_3$}: \textit{If interview nervousness occurs, then interview nervousness makes Patrick's hands shake.}
\item \textbf{$\fml{R}_4$}: \textit{If interview nervousness occurs, then interview nervousness makes Diana's hands shake.}
\item \textbf{$\fml{R}_5$}: \textit{If speech nervousness occurs, then speech nervousness gives Patrick butterflies.}
\item \textbf{$\fml{R}_6$}: \textit{If speech nervousness occurs, then speech nervousness gives Diana butterflies.}
\item \textbf{$\fml{R}_7$}: \textit{If interview nervousness occurs, then interview nervousness gives Patrick butterflies.}
\item \textbf{$\fml{R}_8$}: \textit{If interview nervousness occurs, then interview nervousness gives Diana butterflies.}
\item \textbf{$\fml{F}_1$}: \textit{Speech nervousness occurred.}
\item \textbf{$\fml{F}_2$}: \textit{Interview nervousness occurred.}

\item \textbf{Fact}: \textit{In fact, Patrick's hands did not shake during his speech and he didn't get butterflies.}
\item \textbf{Explanation}: \textit{If individuals have practiced stress-management techniques, then they may not exhibit shaky hands or butterflies in the stomach when nervous.}
\end{itemize}

\end{document}